\documentclass[a4paper, 10pt, conference]{ieeeconf}

\usepackage{tabularx,booktabs}
\usepackage[flushleft]{threeparttable}
\usepackage{nameref}
\usepackage{amsmath}
\usepackage[utf8]{inputenc}
\usepackage{booktabs}
\usepackage{cite}
\usepackage{graphicx}
\usepackage{color}
\usepackage{url}
\usepackage{cite}
\usepackage{comment}
\usepackage{amsfonts}
\usepackage[per-mode=symbol]{siunitx}
\usepackage{hologo}
\usepackage[T1]{fontenc}
\usepackage{aecompl}
\IEEEoverridecommandlockouts          
\begin{document}

\title{\LARGE \bf Impedance Matching: Enabling an RL-Based Running Jump in a Quadruped Robot}

\author{Neil Guan$^{1}$, Shangqun Yu$^{1}$, Shifan Zhu$^{1}$, and Donghyun Kim$^{1}$
\thanks{$^{1}$ University of Massachusetts Amherst, 140 Governors Dr, U.S. {\tt\small donghyunkim@cs.umass.edu}}
}

\maketitle
\thispagestyle{empty}
\pagestyle{empty}

% \author{
% Neil Guan\\
% Department of Electrical and Computer Engineering\\
% University of Massachusetts Amherst
% United States\\
% \texttt{nguan@umass.edu} \\
% %% examples of more authors
% \And
% Shangqun Yu \\
% Manning College of Information & Computer Sciences \\
% University of Massachusetts Amherst \\
% United States
% \texttt{shangqunyu@umass.edu} \\
% \And
% Shifan Zhu \\
% Manning College of Information & Computer Sciences \\
% University of Massachusetts Amherst \\
% United States
% \texttt{shifanzhu@umass.edu} \\
% \And
% Donghyun Kim \\
% Manning College of Information & Computer Sciences \\
% University of Massachusetts Amherst \\
% United States
% \texttt{donghyunkim@cs.umass.edu} \\

%===============================================================================
\begin{abstract}
Replicating the remarkable athleticism seen in animals has long been a challenge in robotics control. Although Reinforcement Learning (RL) has demonstrated significant progress in dynamic legged locomotion control, the substantial sim-to-real gap often hinders the real-world demonstration of truly dynamic movements. We propose a new framework to mitigate this gap through frequency-domain analysis-based \emph{impedance matching} between simulated and real robots. Our framework offers a structured guideline for parameter selection and the range for dynamics randomization in simulation, thus facilitating a safe sim-to-real transfer. The learned policy using our framework enabled jumps across distances of $55~\si{\centi\meter}$ and heights of $38~\si{\centi\meter}$. The results are, to the best of our knowledge, one of the highest and longest running jumps demonstrated by an RL-based control policy in a real quadruped robot. Note that the achieved jumping height is approximately $85\%$ of that obtained from a state-of-the-art trajectory optimization method, which can be seen as the physical limit for the given robot hardware. In addition, our control policy accomplished stable walking at speeds up to $2~\si{\meter\per\second}$ in the forward and backward directions, and $1~\si{\meter\per\second}$ in the sideway direction.
\end{abstract}

% \keywords{Quadruped robots, Running jump, Reinforcement learning} 
%===============================================================================

\section{Introduction}
Unlike wheeled systems, legged robots can navigate irregular terrains by leveraging their multiple limbs. This ability allows them to overcome substantial obstacles and wide gaps by running and leaping across diverse terrains. Numerous studies have sought to enhance the dynamic locomotion of legged robots, with notable demonstrations including high-speed running and hurdling \cite{park2017high}, and performing backflips \cite{8793865}. Unofficially, Boston Dynamics' Atlas robot has shown impressive acrobatic movements across various terrains \cite{atlas_partners}. Despite these advancements, a common limitation is that the controllers and planners are painstakingly calibrated for specific motions or gaits. Furthermore, model-based methods necessitate model simplification \cite{di2018dynamic}, limiting their ability to exploit the full dynamics of the robot.

Reinforcement Learning (RL) emerged as a promising approach for dynamic motion control in legged robots. By effectively exploring the state and action spaces using simulations that encompass full-body dynamics, RL addresses issues tied to limited versatility and model simplification. Advances in RL have led to the development of robust policies capable of navigating challenging terrains in the real world \cite{margolisyang2022rapid,lee2020learning,pmlr-v205-agarwal23a, Miki_2022}. Furthermore, a burgeoning interest in honing more advanced skills in bipedal and quadrupedal robots \cite{pen_2022_ase,li2023robust, RoboImitationPeng20,rudin2022advanced, caluwaerts2023barkour} is opening a new era of RL-based control in legged robotics.

However, there is one fundamental challenge with RL-based dynamic motion control, the so-called sim-to-real issue, and it hampers the deployment of simulation-trained policies in the real world. To address this problem, notable efforts have been made, including dynamics randomization~\cite{rudin2022learning,li2023robust,tan2018sim,pmlr-v205-agarwal23a,doi:10.1126/scirobotics.aau5872,lee2020learning,Miki_2022,margolisyang2022rapid,kumar2021rma,xie2021dynamics,9308468}, meta learning~\cite{yu2020learning,song2020rapidly}, and domain adaptation \cite{RoboImitationPeng20,9308468}. Among these, domain randomization is a frequently used method, but it often lacks a formal guideline for selecting both the randomization variables and the range, even though poorly implemented domain randomization can impair policy effectiveness~\cite{xie2021dynamics}. 

\begin{figure*}
\centering
\includegraphics[width=\linewidth]{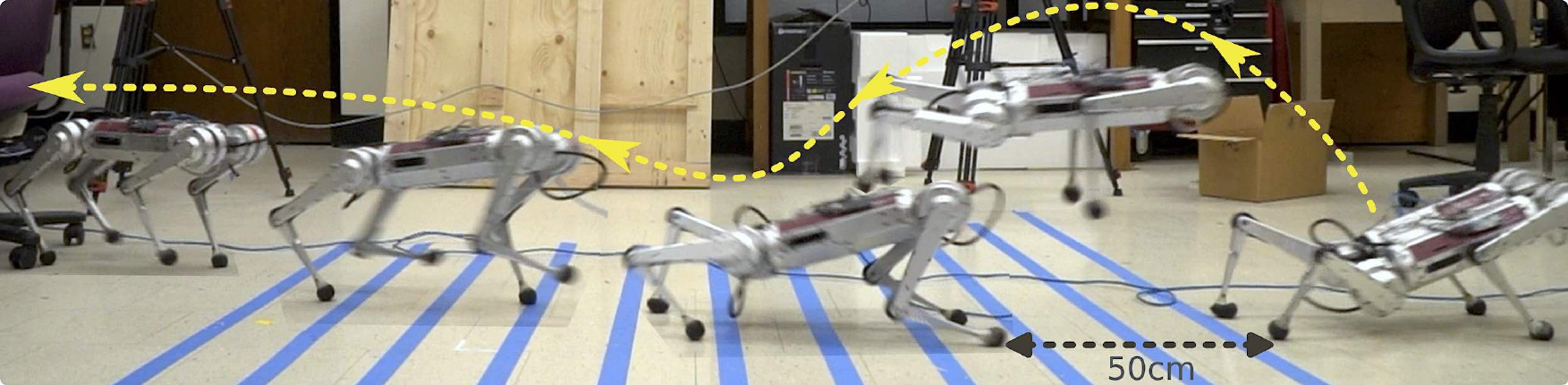}
\caption{{\bf Run and jump demonstration of the Mini-Cheetah robot using our trained policy.} Our novel learning framework enables the quadruped robot to execute dynamic behaviors, such as running and jumping, while significantly narrowing the simulation-to-reality gap. The maximum jumping length and height achieved by our policy are comparable to those obtained through trajectory optimization and the state-of-the art RL-based policies. This suggests that our results are approaching the maximum capability of the given robotic hardware. }
\label{fig:jump_snap}
\end{figure*}

%\dhk{Change all captions in this format: bold title and proper explanation}}

Indeed, we can substantially reduce the gap between simulation and reality by synchronizing the robot's output in both realms, given identical command inputs. For instance, if the errors in joint position from the desired joint positions are consistent between the simulated and real robot, the discrepancy in overall locomotion behaviors should be minimal. Time domain system identification was explored by \cite{caluwaerts2023barkour} to align the actuation parameters between simulation and reality, albeit briefly. Other research focuses on modeling actuators \cite{tan2018sim} or learning actuator dynamics \cite{doi:10.1126/scirobotics.aau5872}, but they overlook physics engine implementation details and neglect modeling outside the actuators, such as belt and link contact friction, which can cause an increase in error magnitude when the commanded motion is rapid. 

This insight forms the foundation of our method: sim-to-real synchronization based on frequency-domain analysis. Instead of naively copying the feedback gains used in the real robot, we conduct a frequency sweep in both real and simulated robots and select appropriate actuator parameters for simulation by matching their bode plots. We refer to our framework as `impedance matching.' While we recognize that our use of \emph{impedance} -- referring here to joint stiffness and damping -- diverges from its technical definition in \cite{hogan1984impedance}, we believe it effectively encapsulates our method's essence. Our framework offers two critical advantages: 1) It allows us to set up a simulation environment that mirrors a robot performing highly dynamic movements, as the frequency sweep can easily cover beyond $20~\si{\hertz}$, wide enough to encompass running and jumping, and 2) it enables us to reduce the randomization range based on experimental data, as we can approximate the true value and variance from multiple, real-world, hardware frequency analysis experiments.

In our RL training, we implemented function preserving transformations, typically used in supervised learning \cite{chen2016net2net}, to maintain normal walking behavior while learning to jump in a curriculum learning setup \cite{doi:10.1126/scirobotics.aau5872,rudin2022learning,kumar2021rma}. Our method centers on developing a jumping behavior derived from an earlier-trained walking behavior. Unlike some research exploring simultaneous learning of multiple behaviors \cite{1802.01561,vollenweider2022advanced} or training an individual policy for each behavior \cite{chen2022humanlevel,caluwaerts2023barkour}, we showed that a jumping behavior can be learned with only minor adjustments to a walking policy.

Our key contributions include: 1) The first introduction of frequency-domain analysis in RL, termed `impedance matching', to reduce the sim-to-real gap; 2) A versatile control policy enabling smooth transitions between dynamic movements such as walking, running, and jumping; 3) The successful real-world deployment of the policy on Mini-Cheetah-Vision, evaluated extensively in dynamic locomotion involving high and long jumping tasks.

\section{Related works}
While there is an abundance of research enabling impressive acrobatic behaviors of robots in simulations~\cite{2018-TOG-deepMimic, Hong_Han_Cho_Shin_Noh_2019, Kwon_Lee_Panne_2020}, actual demonstrations of jumping with real quadruped robots are relatively rare. Notably, \cite{park2017high} showcased an impressive running jump; however, it relied on a planar model for real-time trajectory optimization, limiting its applicability for 3D omnidirectional running jumps. Works such as \cite{matt2021Aeiral} successfully pushed the Mini-Cheetah robot's physical capabilities by combining centroidal dynamics with variational-based optimal control. Yet, the method's offline optimization requirement constrains its capacity for real-time, spontaneous behavior. 

\cite{margolis2021learning} used Depth-based Impulse Control to facilitate the RL-based running jumps over gaps, but at the cost of omnidirectional capabilities. Similarly, \cite{rudin2022advanced} managed to train generalist RL policies across multiple environments, but these policies lacked essential motor skills like walking backward, relying exclusively on forward movement and turning for navigation. Several other studies employed an imitation learning component in their RL approaches, where the policy learns by mimicking a reference motion \cite{li2023robust,smith2023learning,RoboImitationPeng20}. However, the need for reference trajectories limits the scalability of the methods to address various situations including unseen environmental hazards. In contrast, our approach leverages a unique reward design in lieu of imitation learning. This strategy enables us to apply a simulation-trained policy directly in the real world without the need for additional motion retargeting \cite{smith2023learning,RoboImitationPeng20} or manual creation of reference trajectories \cite{li2023robust}.

Recently, \cite{caluwaerts2023barkour} proposed using policy distillation to combine walking, slope climbing, and jumping policies into a single generalist neural network, but they did not demonstrate on-the-fly policy transitions. Our implementation, however, can easily transition between different behaviors, thereby allowing our robot to modulate its speed freely before executing a jump.

%===============================================================================

\section{Sim-to-real synchronization based on frequency-domain analysis}
\label{sec:synch}

To streamline our analysis, we formulate two assumptions. The first one is that the joint dynamics can be closely approximated by a second-order spring-damper system, represented by
\begin{align}
 &I\ddot{\theta} + b\dot{\theta} = \tau \left( = K_p (\theta_{\rm des} - \theta) + K_d (\dot{\theta}_{\rm des} - \dot{\theta}) \right), \\[1mm]
 &I\ddot{\theta} + (K_d + b)\dot{\theta}  + K_p \theta = K_p \theta_{\rm des} + K_d \dot{\theta}_{\rm des},    
\end{align}
where $I$ and $b$ refer to the actuator's inertia and viscous friction, respectively, while $K_p$ and $K_d$ denote the proportional (P) and derivative (D) feedback gains. The second assumption is that the inertia between the actual and simulated robot should be roughly equivalent. %This arises because only $(K_p)/I$ and $(K_d+b)/I$ can be discerned through frequency analysis.% 
To maintain the veracity of this assumption, we disassembled the robot and measured the mass and center of mass (CoM) for each link element by element, and roughly determined our inertia values via CAD models. We also assume that non-viscous and state-dependent friction terms such as static/kinetic/Coulomb friction and load/temperature effects are negligible for high-speed locomotion \cite{b3c2715a826847b9b8b5282c93c1f325}.

\begin{figure*}
\centering
\includegraphics[width=\linewidth]{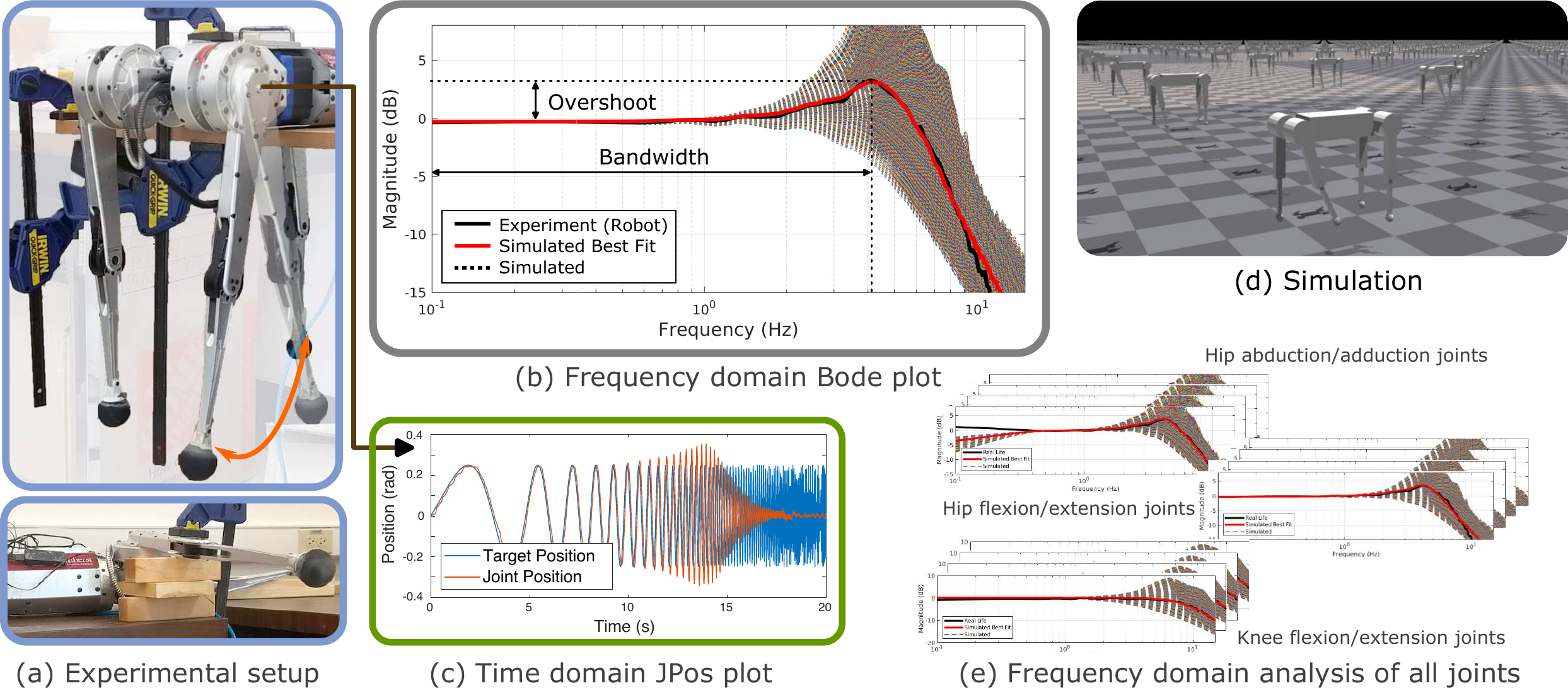}
\caption{Chirp signal test and frequency response analysis of a real and simulated robot}
\label{fig:freq_analysis}
\end{figure*}

Fig.~\ref{fig:freq_analysis} (a) illustrates our experimental setup. We conduct frequency response experiments on each joint individually using a chirp signal, sweeping the joint position command from $0.1~\si{\hertz}$ to $25~\si{\hertz}$ with an amplitude of $0.25~\si{\radian}$. The frequency range was determined based on the Nyquist frequency, which corresponds to the simulation data's sampling frequency of $50~\si{\hertz}$, set by our RL policy update frequency. The chosen amplitude is large enough to yield significant measurements given the motor encoder's resolution while remaining small enough to avoid physical damage from resonance or exceeding the motor's maximum torque. Before performing a sine sweep, we secured the robot's body to a stable platform, employing Proportional-Derivative (PD) feedback control to hold all joint positions at static posture except the joint under examination. When testing knee joints, we secured the thigh link, as depicted in Fig.~\ref{fig:freq_analysis}(a), to avoid the coupled behavior of hip and knee flexion/extension joints. This frequency sweep procedure was reiterated for all joints. We executed the parallel frequency sweep experiment in simulation, as depicted in Fig.\ref{fig:freq_analysis}(c), testing various feedback gains for each actuator across a $50 \times 50$ grid. The values for the proportional gain, $K_p$, ranged from $13 - 27\si{\newton\meter\per\radian}$, while the derivative gains spanned a range of $0.1 - 0.7~\si{\newton\meter\second\per\radian}$.

We employed MATLAB to estimate the transfer function and convert the measured and commanded joint positions in the simulation and real robots into Bode plots. The phase plot was disregarded for two reasons: 1) the time delay, a significant determinant of the phase plot, behaves differently in simulations than in physical systems and measuring exact delay is also difficult, and 2) the remaining information in the data (bandwidth and overshoot) presented in the magnitude plot provide sufficient information to identify the system characteristics of interest. We identified the appropriate gains for use in the simulation by finding the gain pair that yielded the smallest mean-squared error between the Bode plots of the real and simulated systems over the frequencies near the natural frequency. This error computation in proximity to peak magnitude helps identify matches with similar bandwidth and overshoot. Thus, we employed data ranging from $0.1 - 15 \si{\hertz}$ for the knee joints and $1 - 10 \si{\hertz}$ for the hip abduction and flexion joints to find the best match. 

\begin{table}
    \centering
    \begin{tabular}{|l|l|l|} 
    \cline{2-3}
    \multicolumn{1}{l|}{} & $K_p(\si{\newton\meter\per\radian})$ & $K_d (\si{\newton\meter\second\per\radian})$\\ 
    \hline
    Hip roll & 20.6, 18.1, 18.5, 21.0      & 0.492 ,0.516, 0.431, 0.49   \\
    Hip pitch & 16.5, 17.7, 16.9, 18.9      & 0.382, 0.406, 0.382, 0.431   \\
    Knee pitch & 21.8, 22.0, 22.2, 21.8      & 0.541, 0.523, 0.553, 0.553   \\
    \hline
    \end{tabular}
    \caption{Empirically determined gains (FR, FL, HR, HL)}
    \label{tab:gains}
\end{table}

Our choice of feedback gains ($17/0.4$ for $K_p/K_d$) was a reimplementation of \cite{ji2022concurrent}. In theory, the selection can be arbitrary to an extent, as changing the feedback gains will only change the joint target position output from the policy for any given torque output, which is learned via reinforcement learning. We found success in choosing the same gains to get the policy to converge, albeit perhaps choosing a set of critically damped gains would be more stable. 

Upon repeating the search across all twelve joints, we successfully established experimental gain pairs, as delineated in Table~\ref{tab:gains}. We noted a substantial disparity between the feedback gains used in the robot hardware ($17/0.4$ for $K_p/K_d$) and those found in the simulated robots that exhibit similar impedance behavior to the physical system. Domain randomization can address changes in a robot's physical properties due to manufacturing tolerances, calibration or wear-and-tear \cite{9308468}, so we used the empirically determined gain variances in our experiments to roughly obtain the smallest domain randomization ranges encompassing our experimental data. As per Table~\ref{tab:gains}, we set the feedback gains to be used during simulation training: $20, 17.5, 21.5 \si{\newton\meter\per\radian}$ for $K_p$, and $0.45, 0.4, 0.55 \si{\newton\meter\second\per\radian}$ for $K_d$, corresponding to the hip abduction/adduction, hip flexion/extension, and knee flexion/extension joints, respectively. We executed domain randomization in simulated robots by adding values sampled from $\mathbf{U}^{12}(-2.0,2.0)$ and $\mathbf{U}^{12}(-0.05,0.05) \si{\newton\meter\second\per\radian}$.

\begin{figure}
    \centering
    \includegraphics[width=\linewidth]{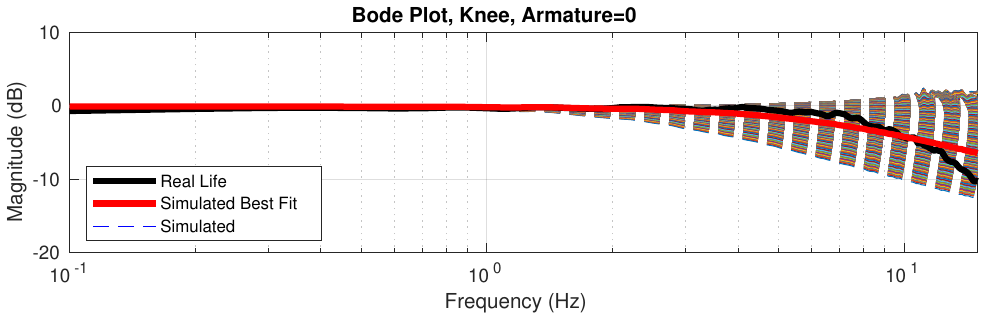}
    \caption{Knee Bode plot assuming zero rotor inertia.}
  \label{fig:armature}
\end{figure}

We demonstrate through impedance matching that high-frequency sim-to-real transfer with robots with geared actuators is very difficult without addressing rotor inertia, which is often not mentioned in various RL locomotion papers. Isaac Gym enables rotor inertia to be approximately simulated via its rigid body armature paramater, which is a constant array added to the diagonals of the mass matrix of the robot, a critical element in fitting Bode plots from simulation to empirical measurements. Assuming rotor inertia is zero, the best-fit simulation parameters do not match real life dynamics, as shown in Fig.~\ref{fig:armature}, as our grid-search across a wide range of gains yields no match to real life. The rotor inertia markedly influences the best fitted gains; thus, careful selection is crucial. We used $0.000072  \times 9.33^2 (\simeq 0.0063) \si{\kilogram}$ for the reflected rotor inertia of knee joints and $0.000072  \times 6^2 (\simeq 0.0026) \si{\kilogram}$ for the other joints. Note that we incorporated detailed actuator models ~\cite{katz2018low} into Isaac Gym, which include aspects such as rotor inertia, the back-emf effect, dry friction, and bus (battery) voltage limits, which were experimentally measured using a custom dynamometer. These additions allow us to accurately simulate actuator behavior, particularly at the edge of its operational conditions. 
    
%=============================================================================== 

\begin{figure*}
\centering
\includegraphics[width=\linewidth]{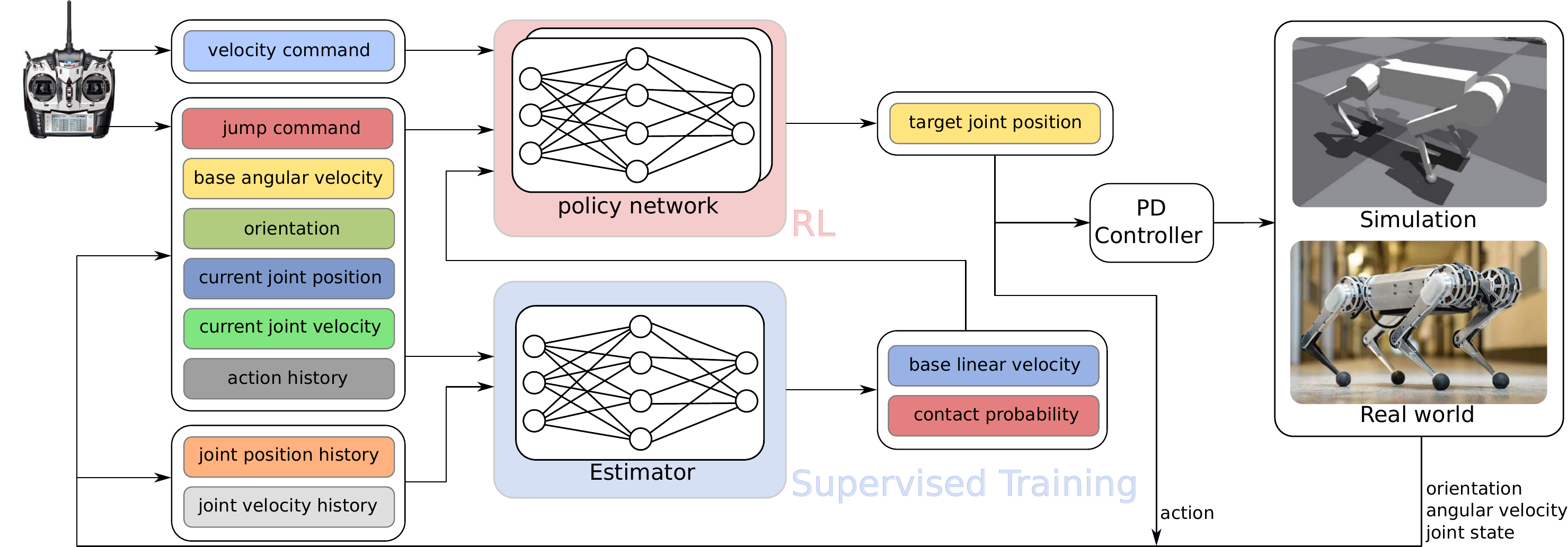}
\caption{Main pipeline of the RL agent}
\label{fig:arch}
\end{figure*}

\section{Training framework to learn walking, running, and jumping}
\label{sec:training}
% Net2Net presents the idea of instantaneously transferring the knowledge of a network to a deeper or wider network to accelerate the training of a significantly larger network through function preserving transformations~\cite{chen2016net2net}. 

In order to acquire multiple skills for a single RL agent, we adopted the idea of Net2Net \cite{chen2016net2net} for multi-task RL. In particular, we use a modified Net2WiderNet implementation performed on the command inputs of a network. This is done by adding a blank neuron by adding a row of zeros to the input layer weight matrix. This blank neuron can be trained to freely signal the transition into and out of the new behavior, trained with conditional rewards. To signal a jump, the blank neuron begins at 0 to walk, transitioning to 1 in the jumping state, and returning to 0 to land. We train in multiple phases. During Phase 1, we train the agent how to walk. During Phase 2, we add a jump command input neuron to the model's input layer and train both walking and jumping in a 50/50 split to avoid catastrophic forgetting of walking. Notably, the jump can be controlled using velocity commands just as walking could. To our knowledge, this is the first time Net2Net has been used for locomotion.

Our reward terms take inspiration from \cite{rudin2022learning,kumar2021rma,rudin2022advanced,ji2022concurrent}. Additional rewards introduced starting in Phase 2 are shown in Table~\ref{tab:Reward}. Of note is the dense jump reward, which shapes the jumping behavior. This simplistic jump reward can train a jump without requiring a reference motion. The sparse jump is intended to further shape the jump to be large enough to cross gaps. Phase 2 is split into subphases 2a and 2b. The dense jump reward encourages jumping motions, but should be lowered after the robot is capable of jumping. The sparse jump reward encourages consistent jumps based on the desired liftoff velocity, which we set at $2.5 \si{\meter\per\second}$. 
% \begin{table}[]
%     \centering
%     \begin{tabular}{ |p{4cm}|p{5cm}|p{1cm}|p{1cm}|  }
%         \hline
%         \multicolumn{4}{|c|}{Rewards} \\
%         \hline
%         Reward Term& Expression &Reward Scale (Walk)&Reward Scale (Jump)\\
%         \hline
%         \multicolumn{4}{|c|}{Shaping Rewards} \\
%         \hline
%         Action Magnitude   & ${||q_t-q_{nominal}||}^2$    &$-5.0$&   $0.0$\\
%         Vertical Linear Velocity&   ${||V_z||}^2$  & $-50$   &$0.0$\\
%         Smoothness &${||\tau_t-\tau_{t-1}||}^2$ & $-0.2$&  $-0.2$\\
%         Orientation    &${||V_z||}^2$ & $-20$&  $-120$\\
%         Collision&   $1$  & $-50$&$-50$\\
%         Joint Torque Limits& $\sum{relu(\tau-\tau_{limit})}$  & $-900$&$-900$\\
%         Joint Position Limits& $-relu\left(-q_t+q_{lower\ limit}\right)+relu\left(q_t-q_{upper\ limit}\right)$  & $-150$&$-150$\\
%         Foot Acceleration& $||v_{feet,\ t}-v_{feet,t-1}||$  & $-0.5$&$-0.5$\\
%         Foot Clearance& $\sum{\left(0.09-z_{feet}\right)^2\ast\sqrt{v_{feet}}}$  & $-250$   &$-250$\\
%         Termination& $1$  & $-500$&$-500$\\
%         Torque& ${||\tau||}^2$  & $-0.08$&$-0.08$\\
%         \hline
%         \multicolumn{4}{|c|}{Task Rewards} \\
%         \hline
%         Tracking Linear Velocity& $\exp(-0.25||cmd_{v_{xy}}-V_{xy}||)$  & $45$&$45$\\
%         Tracking Angular Velocity& $\exp(-0.25||cmd_\omega-\omega||)$  & $30$&$30$\\

%         \hline
%     \end{tabular}
%     \caption{Reward Terms}
%     \label{tab:Reward Terms}
% \end{table}

\begin{table*}
    \centering
    \caption{Jump Reward Terms}
    \label{tab:Reward}
    \begin{tabular}{|l|l|l|l|l|} 
    \hline
    Reward Term & Expression  & Scale, (Phase 1) & Scale(Phase 2a)& Scale (Phase 2b)  \\ 
    \hline
    Dense Jump  & $-std(F_{foot})$ & $0.0$ & $-2.5$ & $-0.25$ \\
    Sparse Jump & $exp(-{(V_{liftoff}-V_{liftoff,desired})}^2/2)$ & $0.0$ & $250$ & $250$ \\
    \hline
    \end{tabular}
\end{table*}

In multi-task RL, it is often challenging for a single policy to acquire distinct skills due to the presence of opposite gradient directions and nonstationary data. In our case, we note that training both walking and jumping simultaneously from scratch does not converge to distinct behaviors, whereas if we train them sequentially, we obtain distinct behaviors, leading us to believe that the gradients corresponding to one task oppose the gradients corresponding to the other. Segmenting different behaviors into different states not only grants the freedom to engineer different behaviors individually but also grants interpretability in allowing a higher level policy to choose which behaviors are best. 

%===============================================================================

\subsection{System setup:} We use the MIT Mini Cheetah Vision as our experimental platform. The robot stands $30~\si{\centi\meter}$ tall and weighs $12~\si{\kilo\gram}$. The robot features 12 custom-designed proprioceptive actuators, each with a maximum output torque of $17~\si{\newton\meter}$. Our model has a slightly longer body length compared to the original MIT Mini Cheetah~\cite{8793865}. Our robot uses a minimal sensing suite, allowing our framework to be generalizable to many robots, including relatively inexpensive ones. We use data from the robot’s accelerometer, gyroscope and joint encoders for the robot’s walking, running and jumping capabilities. We use Isaac Gym \cite{rudin2022learning} to train the quadrupedal robot and adapted the code from \cite{ji2022concurrent} for the sim-to-real deployment. 

\subsection{Estimator network:} We follow \cite{ji2022concurrent} in our estimator network implementation on the mini cheetah. Our estimator network takes in the base angular velocity, the projected gravity, the joint position history, the joint velocity history, the action history and the jumping mode. The joint position histories are sampled at $t$, $t-0.02$, and $t-0.04$, and the action history is sampled at $t-0.02$ and $t-0.04$. The estimator network predicts foot contact probability and the estimated linear velocity of the robot and is trained via supervised learning. The estimator network is implemented with a $[256 \times 128]$ structure.

\subsection{Control Policy:} The control policy is trained with PPO~\cite{schulman2017proximal}, with both actor and critic networks having a $[1024 \times 512 \times 256]$ structure. The observation space consists of the estimated velocity, the foot contact probabilities, the base angular velocity, the projected gravity, the $x$, $y$, yaw velocity commands, the jump command, the joint position, the joint velocity, and the previous actions. The sensing and inference update loop runs at $50~\si{\hertz}$, while the actuator-level PD controller runs at $40~\si{\kilo\hertz}$. The action is a set of target joint position vectors that will be assigned to the PD controller. We sampled $x$ velocity commands from $\mathbf{U}^1(-3.0, 3.0) ~\si{\meter\per\second}$, $y$ velocity commands from $\mathbf{U}^1(-1.0, 1.0) ~\si{\meter\per\second}$, and yaw angular velocity commands from $\mathbf{U}^1(-2.0, 2.0)~\si{\radian\per\second}$. Jump commands are randomly sampled between $0$ and $1$.  

\subsection{Domain Randomization:} Domain randomization is employed to prevent overfitting to the simulation dynamics. Decimation is randomized to $\mathbf{U}^1(-2,2) \si{\second}$ to mitigate the effect of variable system delay on the real robot. Ground friction is randomized to $\mathbf{U}^1(0.05,3.0) \si{\newton\meter\per\radian}$. We randomize stiffness and damping by adding a value sampled from $\mathbf{U}^{12}(-2.0,2.0)$ and $\mathbf{U}^{12}(-0.05,0.05) \si{\newton\meter\second\per\radian}$ to our commanded stiffness and damping, respectively, based off our impedance analysis. Shank length is randomized from $\mathbf{U}^4(0.18,0.20) \si{\newton\meter}$ to help mitigate the effects of foot deformation. Base mass is randomized from $\mathbf{U}^1(-0.4,1.6) \si{\kilogram}$. Joint position noise is sampled from $\mathbf{U}^{12}(-0.05,0.05) \si{\radian}$ to account for poor encoder measurements and deviations in the zero position of the motor. Joint velocity noise is sampled from $\mathbf{U}^{12}(-0.5,0.5) \si{\radian\per\second}$. Angular velocity noise is sampled from $\mathbf{U}^3(-0.2,0.2) \si{\radian\per\second}$. Projected gravity noise is sampled from $\mathbf{U}^3(-0.05,0.05)$

%===============================================================================

\section{Experimental Results}
\label{sec:result}

\begin{figure}  
    \centering
    \includegraphics[width=\linewidth]{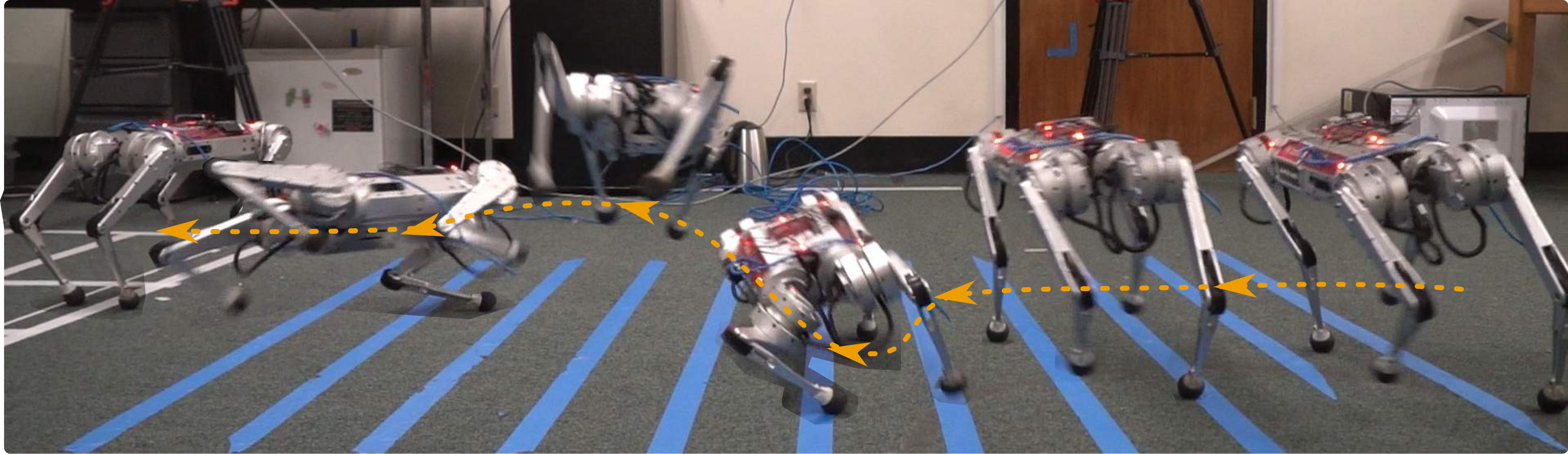}
    \caption{A sideways running jump using naive policy is not stable. }
  \label{fig:naivepolicyinstability}
\end{figure}

%\dhk{remove the last stand up robot}

To prepare a consistent way to test each jump, we start the robot from a standstill and write a short script to accelerate to a specified velocity, jump for a specified time interval, and decelerate to a stop for a total of five jumps for each velocity. We perform this experiment with two trials, one trained with our experimentally determined gains, and another trained with the naive gains. Table~\ref{tab:heightdistance} lists the mean and standard deviation jump height and distance of the robot at various speeds for both trials. 

The data is obtained via a motion capture rig. We measure the difference between the minimum and maximum height of the tracker when the robot jumps to approximate the jump height, and we find the distance by taking the difference between the robot's X-Y position during the inflection points of the jump. Although the naive method sometimes matches or even exceeds the distance with the matched impedance method, the average jump distance standard deviation is $230\%$ that of the matched impedance policy. Similarly, the jump height average standard deviation is $148\%$ that of the matched impedance policy, further indicating inconsistent behavior in the naively trained policy. Fig~\ref{fig:naivepolicyinstability} illustrates an unstable but long sideways jump from the naive policy, which in our experience, was much more present in the naive policy. We attribute the inconsistency and instability to the poorer sim-to-real transfer in the naive method. A stable sim-to-real transfer is a consistent sim-to-real transfer, and the impedance matched trials' lower variance in jump height and distance indicate a better sim-to-real transfer.

\begin{table*}
    \centering
    \begin{tabular}{ |p{3.0cm}|p{0.85cm}|p{0.85cm}|p{0.85cm}|p{0.85cm}|p{0.85cm}|p{0.85cm}|p{0.85cm}|p{0.85cm}|  }
        \hline
        Command & 
        \multicolumn{4}{c|}{Jump Height (m)} &
        \multicolumn{4}{c|}{Jump Distance (m)} \\
        \cline{2-9}
        & \multicolumn{2}{c|}{Naive} & \multicolumn{2}{c|}{Matched} &\multicolumn{2}{c|}{Naive} & \multicolumn{2}{c|}{Matched} \\
        \cline{2-9}
         & Mean & Std & Mean & Std & Mean & Std & Mean & Std\\
        \hline
        Forwards, 1.0 m/s & 0.332 & 0.008 & 0.318 & 0.014 & 0.690 & 0.076 & 0.597 & 0.042\\
        Forwards, 2.0 m/s & 0.246 & 0.022 & 0.244 & 0.013 & 0.944 & 0.030 & 0.969 & 0.026\\
        Sideways, -1.0 m/s & 0.304 & 0.061 & 0.374 & 0.016 & 0.636 & 0.092 & 0.587 & 0.054\\
        Sideways, 1.0 m/s & 0.249 & 0.041 & 0.366 & 0.052 & 0.701 & 0.183 & 0.542 & 0.051\\
        Backwards, -2.0 m/s & 0.346 & 0.031 & 0.381 & 0.015 & 0.944 & 0.030 & 0.932 & 0.025\\
        \hline
        
    \end{tabular}
    \caption{Real-life Jump Height and Distance}
    \label{tab:heightdistance}
\end{table*}

\cite{2110.06330} provides a highly optimized upper bound on the jump height of the Mini-Cheetah, achieving jumps onto $34~\si{\centi\meter}$ platforms, roughly corresponding to approximately $45~\si{\centi\meter}$ jump heights. Treating their work as the robot's physical limit, we were able to achieve $85\%$ of this height consistently while performing backward running jumps. 

% By utilizing reinforcement learning, we were able to perform running jumps at speeds of up to $-2.5 \si{\meter\per\second}$, which is $90\%$ faster than the fastest reference trajectory generated by \cite{2110.06330}. 

We place tape markers on the ground every 20 cm denoting the robot’s distance jumped and go frame-by-frame on the video to determine where the robot entered and exited its airborne phase to roughly estimate the maximum gap each jump can cover. In all trials, the maximum gap the robot can cross is $55~\si{\centi\meter}$ going forward at $2~\si{\meter\per\second}$, $50~\si{\centi\meter}$ going backward at $2~\si{\meter\per\second}$, and $40~\si{\centi\meter}$ going sideways. 

\begin{figure}
    \centering
    \includegraphics[width=0.92\linewidth]{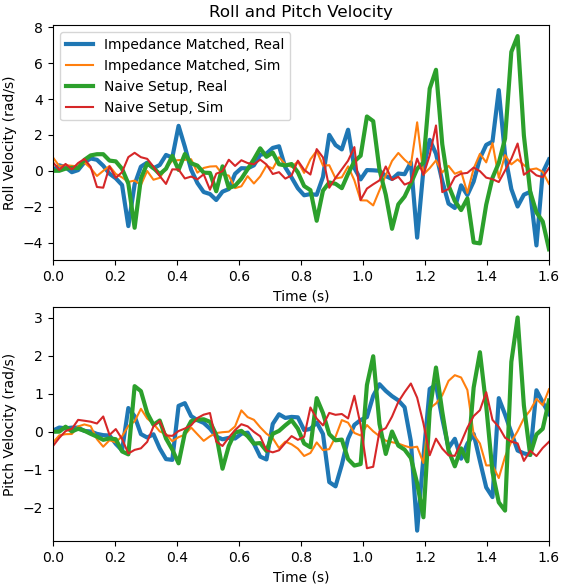}
    \caption{Angular velocities (roll, pitch) of a robot under the $2.25~\si{\meter\per\second}$ forward running command.}
  \label{fig:accelerationrollpitch}
\end{figure}

We record the roll and pitch velocity readings during the acceleration phase for both the naive and impedance matched policy, as shown in Fig~\ref{fig:accelerationrollpitch}. 

%===============================================================================

\section{Limitations}
\label{sec:limitations}

Our method assumes good inertia estimations, which were determined using CAD models and are difficult to verify. The same actuators performing similar roles have vastly different experimentally determined gains. Although factors such as the simulation implementation, belt friction and link contact friction may account for some of these differences, it is unclear how much these differences are from inaccurate inertia estimations, and as such, applied loads on the robot such as ground forces cause unmodeled sim-to-real variables. Alternative approaches \cite{tan2018sim} can be to assume uniform mass distribution for each link, or perform domain randomization on inertia estimations, but even in those contexts, they still depend on an initial inertia estimation, which are often not verified experimentally. Future work should use calibrated mass property measurement systems for empirical inertia verification.
%Likewise, our method assumes the negligible higher-order friction terms. Future experiments can assess its importance by testing impedance matching with higher amplitudes and different periodic target trajectories, and if impedance matching does not match at different amplitudes, then the grid search can be expanded to include other variables.%

We trained running and jumping with random state transitions leads policies, but they learned conservative behaviors that are always ready to transition between each other. For instance, at higher forward speeds, the walking policy turns into a bounding policy, which can easily transition into jumps, and additional reward turning is necessary to prevent the walking policy from changing. Last, pushing the limits of the hardware is not enough to clear meaningful gaps with modern actuators; another necessity is for the robot to step precisely around gaps, which is not addressed by our method. 

%===============================================================================

\section{Conclusion}
\label{sec:conclusion}

We develop a new learning framework to enable highly dynamic maneuvering of a quadrupedal robot including stable walking, rapid running, and high/long jumping via RL. RL is a promising approach for training such controllers, but deploying the learned control policy in real life is notoriously difficult due to the gap between simulation and real life. We propose a method to minimize the sim-to-real gap through frequency-domain analysis with little physical risk to the robot. We provide a jumping controller enabling running omnidirectional jumps, trained with a novel, model-agnostic method. We show that our method enables jumps of a distance of $96~\si{\centi\meter}$ for the robot's center of mass which corresponds to a $55~\si{\centi\meter}$ gap and heights of $38~\si{\centi\meter}$, which is competitive with state-of-the-art performance in quadrupedal robot jumping in real-world tests in distance and height. 

%===============================================================================
\section*{ACKNOWLEDGMENT}
We express our gratitude to Naver Labs and MIT Biomimetic Robotics Lab for providing the Mini-cheetah robot as a research platform for conducting dynamic motion studies on legged robots.
%===============================================================================

\bibliographystyle{IEEEtran} % use IEEEtran.bst style
\bibliography{RL}
% \bibliography{./IEEEabrv,./RL}

\end{document}